# Co-Design and implementation of an open-source 3D printed robot

Cristina Gena, Chiara Vaudano, Davide Cellie
University of Turin

## INSPIRATION

During the 2017-18 academic year we carried out a series of coding activities, lasting about 3 months, in a third of the Giulia Falletti primary school in Barolo in Turin (Gena et a., 2020). These activities aimed to teach students not only the basics of programming, but also to introduce a new language and a new way of thinking and solving problems: computational thinking.

The class consisted of 25 pupils: 14 males and 11 females, which we then operationally divided into two working groups (13 + 12) to make coding lessons more manageable and provide better childcare. The lessons lasted one hour and were conducted, in the presence of one of the teachers, by a computer teacher assisted by a student / facilitator.

The first lesson was preparatory to the presentation of the main topics in the field of *coding:* sequential instructions and algorithms. The contents of the first and second lessons of Course 2 of code.org were then shown to the pupils [1]. In the following weeks the following lessons were covered. Observing the coding exercises closely, we noticed that the children encountered some difficulties, mostly related to problems of spatial orientation (*orienteering*): for example it was still difficult for them to distinguish left and right from the perspective of the virtual robot., especially when the robot on the screen was not turned from their point of view.

At the end of the coding activities, we distributed a satisfaction survey to the children which consisted of 15 questions that the children could answer by giving a rating inserted in a scale of five values expressed thanks to a *smileyometer* (Sim, 2012; Cena et al. 2017; Jameson et al. 2011), which is the most used tool for measuring the opinion of children and includes a rating on a scale where the smileys correspond to a range from 1 to 5 (strongly disagree to strongly agree). We asked the children to express their opinion and preferences by choosing one of the faces.

The results obtained allowed us to conclude that the lessons held were appreciated and satisfied the children: more than 68% of the ratings obtained the degree of satisfaction " *Very much* ", with 84% of the children finding "*interesting the proposed activities* "And 92% who would like " *to have more lessons of this type* ".

At the end of the three months of this positive experience, we realized that having an educational robot that can perform the same kind of actions that virtual robots do, like those of code.org, is very useful for children, especially to help them to solve orientation problems.

Therefore, since no commercial robot had the characteristics we wanted, we decided to create an educational robot from scratch, equipped with social, interactive and emotional skills, able to involve children and establish an emotional bond with them in order to increase their learning and involvement. We decided from the beginning to design the robot as an *open source project,* made at a low cost, proposing a kit that can be easily reproduced and improved by anyone who wishes.

## THE IMPORTANCE OF CO-DESIGN METHODOLOGY

For the specific creation and design of the appearance and structure of the robot and for the definition of its personality and characteristics, we decided to follow a co- design methodology with elementary school children according to the paradigm of cooperation and participatory design.

Creativity and imagination are two of the activities that develop the most during a co-design session, a methodology applied for educational purposes that involves children in the design and creation of prototypes, for the purpose of a final product.

---

[1] Available at: https://studio.code.org/s/course2/

Thanks to their imagination it is possible, for example, to develop prototypes of educational robots capable of being and doing anything, as far as possible, and even if the target may be very young, in every project it is important to bring them to be real designers.

The work can also prove to be satisfactory both for the number of interesting ideas that arise and on which we can work, and for the participation of students, curious and with a great desire to apply themselves in the best way. The desire to contribute is certainly a strong point of this methodology, which underlines and favors it.

In the 2017-2018 academic year, we conducted a co-design session with the same class with which we had done the coding lessons (Cietto et al. 2018). Divided into four groups, each was joined by a university student from the three-year course of Computer Science or Social Innovation, Communication and New Technologies (ICT) of the University of Turin, with the function of moderator and facilitator.

For the conduct of the session we were inspired by the approach to co-design with children by Mechelen and Vanden Abeele [2], freely adapted where required. The methodology we were inspired by allows each child to participate without entering into competition, focusing on teamwork and the final goal. We carried out the co-design session following these steps:

- We explained to the children that their work was important and that they were part of the design team with us and not just end users;
- We have created 4 groups, as heterogeneous as possible (asking for help from teachers);
- 4 adult facilitators (one per group, university students involved in the projects) and 2 coordinators supervised all groups and took notes;
- We had the children draw lots of pictures and then each facilitator helped their group to derive the main characteristics of the robot in the form of keywords or short phrases;
- The facilitators followed the work of the children, coordinated them with moderation and took note of how their discussion unfolded.

Specifically, following the indications of Mechelen and Vanden Abeele, the specific phases in which we organized the practical co-design process were the following:

- *phase 1:* introduction to the project and presentation of the people involved;
- *phase 2:* introduction to the methodology and its rules and design constraints. We have provided some general guidelines on robot implementation requirements and constraints, such as the fact that:
    - the robot should have social skills
    - should be programmable by children
    - could act as an assistant helping them learn programming
    - should have a medium size, more or less as large as a small dog
    - should have a toy style, not similar to a human or an animal
    - could have lights and a screen
    - could emit any sound and voice
    - could speak
    - could be moved on wheels.
- *phase 3:* first phase of conception in which we asked the children, divided into 4 groups, to write all their ideas on post-its. Then with the help of the facilitator, each group had to converge on common factors. In this phase, the children individually drew on a piece of paper the idea of a robot that resided in their mind, and we asked them not to be influenced by images or photographs of existing examples, and also to keep track of all of them on paper. those functionalities and characteristics to add to the personal drawing, but not representable (parts of the robot that are mobile rather than luminous, facial reactions to different moods, dialogues or sound parts, etc.). The role of the examiner was to view and encourage the children as much as possible to indulge themselves as they saw fit, without constraints and fears. It is legitimate for them to propose ideas and particularities on which to begin to familiarize themselves with their own design without, however, influencing them, inducing them towards a specific detail. Some children started straight away, producing a complete idea in a few minutes, unlike others who needed more time to think and reason, then probably creating a cleaner and more detailed drawing. Waiting for all participants to finish is essential to maintain organization and order with the methodology designed;

---
[2] Mechelen, Maarten Van and Vero Vanden Abeele. "Co-design revisited: exploring problematic co-design dynamics in kids" Workshop on Methods of Working with Teenagers in Interaction Design, CHI 2013

- *phase 4:* second phase of conception in which each group had to elaborate its proposal through drawings and storyboards. In this phase we tried to make the various drawings converge in a single ending (Fig. 1), choosing the best physical part of the robot among the various sketches made (eyes, mouth, body, etc.). The choice took place in total collaboration with the boys and, above all, making sure that there was a little of each of them within the finished figure. After examining the various proposals, the final version took shape on a new white sheet. Each necessary part to be designed was chosen by discussing those that worked best in the various initial prototypes and, by voting, reproduced. To conclude this phase, the design was colored by the whole team after a careful and voted choice of the most desired colors, and a name was selected to assign to the finished robot;
- *phase 5:* presentation of the projects and final discussion. The conclusion of the co-design session was the presentation of the finished work where a representative from each team told the character born of their free imagination in front of the whole class, preparing the speech a few minutes before and exposing, mainly, the particularities of creation.

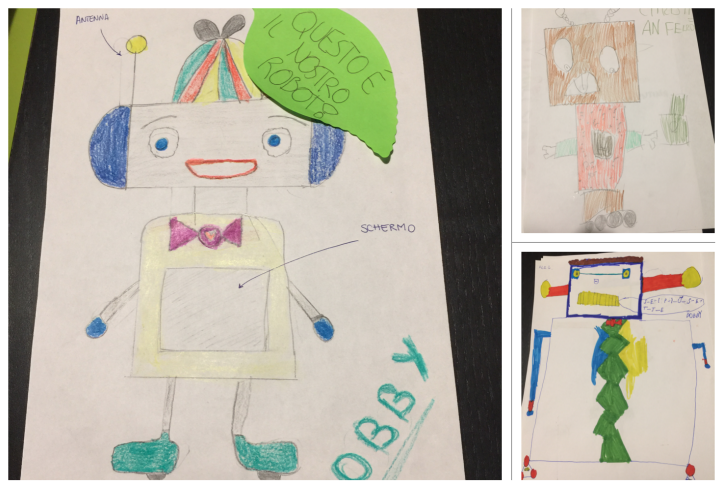

Figure 1. The final drawings of phase 4.

Finally, the children proposed a series of names that were then voted on by the class and the teachers at the end of the session, leading to baptize the future robot with the name of Wolly.
The overall co-design activity lasted for two hours.
The subsequent analysis of all the materials produced during the session, (notes, video of the final presentations, post-its, drawing, field observations, etc.) was inspired by the Grounded Theory (Strauss and Corbin, 1990). defined, inspired by the 'interpretative paradigm' which interprets the processes underlying a given phenomenon, and, in this case, emphasizes the generation of theory from the analysis of qualitative and quantitative data.

In this case, particular reference was made to the first two phases of this methodology: that of the so-called open *coding,* through which the first concepts are identified among the still fragmented data, but from which it is possible to derive the properties and the first categories, still open and flexible; and that of axial *coding,* in which the paradigmatic model is used to establish links between categories and subcategories, so as to identify relationships between the concepts that emerged in the first phase.

From this process it was possible to draw numerous design indications about the expectations of children, among which we can list the most significant:

- The robot is supposed to be a child teacher / assistant who teaches and plays with them and may even be a bit stern at times. It is not just a friend, but it educates in a fun way;
- it should express (positive) emotions thanks to its voice, its cartoon face (eg big eyes and smiley face, etc.) and its bright physical parts. Note that in most of the drawings the nose was absent while the presence of the mouth is considered important to emphasize the emotion behind the interaction;
- The robot could express its emotions through lights. Hence the idea of using a slightly transparent material for the body with small diffused light LEDs inside;
- most of the children, all but one, imagined the robot as male or gender neutral;
- Its body is supposed to be a bit boxy, and several groups have designed playful and colorful clothes / accessories inherent to the character envisioned for the robot. Hence the idea of integrating the final prototype with one of those proposed by the children (hat, bow, cape, etc.), and print the body dressed in jacket and bow tie
- Some children have proposed the use of various useful supports to make him a "handyman": a light, a jet of water, a radio, a camera, a fan to heat, etc;
- The robot should have arms, which we have imagined as passive and replaceable.

**Realization of the robot**
Following the directions of the children, our project team then moved on to the implementation phase. We used a very common robotic kit: an Arduino Mega containing the firmware, an Adafruit controller for the motor, a Powerbank battery and a WeMos D1 Mini ESP8266 board for wireless communication, an Android phone to manage voice interaction and facial expressions. For movement, the robot has been equipped with four independent motorized wheels and employs differential-wheel steering to move. Instead, what allows to control its movements is a web application (will be described later), which uses a series of Rest commands launched to the http server inside the robot as input and which, therefore, allows a simple and intuitive use of its potential.

Before starting with the integration with coding exercises, to test its interactive skills, we implemented the famous Taboo game in Wolly, which proved to be a great way to show children a real example of exchanging communications with the robot. Through play, the children understood even more how to communicate with a machine and how the machine wishes to receive instructions. We made the game like this (Trainito, 2019):

**Rules of the game:**
- the robot provides clues to be able to make children guess a certain word (eg: chair, rain, panda, oceans, etc.);
- for each wrong word the robot provides a new (more intuitive) clue;
- there are a maximum of 3-4 clues per word;
- before giving the answer, the child must wait for the acoustic sound (beep) and at that point the robot listens;

- when the children guess the word will have to answer "yes" or "no" to the question to continue playing

**Playing the game:**
- the game starts with the robot explaining the rules;
- robot provides the first clue and lets the children think about twenty seconds when they play in groups;
- after 20 seconds the robot emits a beep;
- a child of the group says the correct word;
- if it is right robot compliments and is happy, and will ask if the children want to play again;
- if it is wrong, the robot tells it to try again and gives another clue. Continue up to a maximum of 3 or 4 clues (the clues will be more and more intuitive and the last one will be evident).
- if children do not guess, the robot will be sad and comfort them by telling them not to give up and try again. It will also say the word not found.

**Emotions shown by the robot:**
- if children guess at the first clue: very happy robot, eyes to heart.
- if children guess at the second clue: happy robot.
- if children guess at the third / fourth clue: happy robot.
- if children do not guess: sad and disappointed robot.
- between wrong attempts: robot remains neutral.

We experimented with the Taboo game in two field trials (with the same co-design class and with a fifth from another institute). The tests lasted less than an hour, but nevertheless allowed us to formulate the following qualitative observations:
- the children were thrilled to play with a robot. Through play, children understand how to give the correct instructions, for example by speaking clearly and slowly;
- The children enjoyed being all together in a different way and having to reflect to find the common solution (team work);
- The children were taught to make little noise while interacting with the robot and this gave them a sense of responsibility and complicity;
- when the children guessed the word they were delighted and jumped;
- the children were impressed and admired by Wolly, and by its ability to show expressions.

**Thanks**
We would like to thank the 3A class (2018/2018) of the Giulia Falletti Primary School of Barolo in Turin and their teachers Linda Garofano and Teresa Carretta. We thank the ICT students who helped us in the various co-design experiments and the students of the Mazzarello Institute in Turin. Finally, we thank Valerio Cietto and Gianluca Perosino for the first release of Wolly, Mauro Giraudo for 3D printing, and Massimo Trainito for the initial coding lessons.